\documentclass[preprint,12pt]{elsarticle}
\usepackage{graphicx}
\usepackage{subfigure}
\usepackage{amsmath}
\usepackage{ifluatex,ifxetex}
\ifluatex
  \usepackage{fontspec} % optional
\else\ifxetex
  \usepackage{fontspec} % optional
\else % assume that pdftex engine is in use
  \usepackage[T1]{fontenc}
  \usepackage[utf8]{inputenc} % optional for "\ng"
\fi\fi

\usepackage{amssymb}
\biboptions{numbers,compress}
\journal{Nuclear Physics B}
\begin{document}

\begin{frontmatter}



\title{A Hierarchical User Intention-Habit Extract Network for Credit Loan Overdue Risk Detection}

\author[addr1]{Hao Guo\corref{mycorrespondingauthor}}
\ead{guohao@trusfort.com}

\author[addr1]{Xintao Ren}
\ead{renxintao@trusfort.com}

\author[addr1]{Rongrong Wang}
\ead{wangrongrong@trusfort.com}

\author[addr1]{Zhun Cai}
\ead{caizhun@trusfort.com}

\author[addr2]{Kai Shuang}
\ead{shuangk@bupt.edu.cn}

\author[addr1]{Yue Sun}
\ead{sunyue@trusfort.com}

\cortext[mycorrespondingauthor]{Corresponding author(Both Hao Guo and Xintao Ren are corresponding author)}

\address[addr1]{Beijing Trusfort Technology Co.,Ltd., 100089, Beijing, P.R.China}
\address[addr2]{State Key Laboratory of Networking \& Switching Technology, Beijing University of Posts and Telecommunications, 100876, Beijing, P.R.China}

\begin{abstract}
 More personal consumer loan products are emerging in mobile banking APP. For ease of use, application process is always simple, which means that few application information is requested for user to fill when applying for a loan, which is not conducive to construct users’ credit profile. Thus, the simple application process brings huge challenges to the overdue risk detection, as higher overdue rate will result in greater economic losses to the bank. In this paper, we propose a model named HUIHEN (Hierarchical User Intention-Habit Extract Network) that leverages the users’ behavior information in mobile banking APP. Due to the diversity of users’ behaviors, we divide behavior sequences into sessions according to the time interval, and use the field-aware method to extract the intra-field information of behaviors. Then, we propose a hierarchical network composed of time-aware GRU and user-item-aware GRU to capture users’ short-term intentions and users’ long-term habits, which can be regarded as a supplement to user profile. The proposed model can improve the accuracy without increasing the complexity of the original online application process. Experimental results demonstrate the superiority of HUIHEN and show that HUIHEN outperforms other state-of-art models on all datasets.
\end{abstract}



\end{frontmatter}

%\linenumbers

\section{Introduction}
With the progress of mobile Internet, mobile banking APP has become an important platform for banks to provide online services. There are more personal consumer loan products on the platform, where consumers can apply for credit loan business \cite{ref1, ref2}. As credit loans are the cornerstone of the banking industry, and credit loan risk accounts for 60\% of the bank’s overall threats \cite{ref3}, credit risk decision-making is a key determinant of the success of financial institutions \cite{ref4}. If the overdue risk cannot be effectively identified, not only the bad debt rate will increase significantly, causing a great economic loss to the bank, but also it will significantly increase bank management costs. From the perspective of bank security, credit loan risk detection has become an important task for extensive research in the banking industry \cite{ref5, ref6}, where the vital issue is to distinguish between good and bad credit applicants. 
\par
In the mobile banking APP, the consumer loans occupy a large proportion, whose application process is usually simple and fast. To improve user experience and enhance the competitiveness of the loan services, users are not requested to fill in too much personal information, which will reduce the information that the credit loan risk detection system can refer to which is not conducive to control the overdue rate. Therefore, how to detect the credit loan risk and predict the users' online overdue rates to the greatest extent without increasing the complexity of online application process is an important challenge faced by many banks that provide online credit loan services.
\par
The general methods almost rely on statistical strategy in credit loan risk detection areas, which mainly use time-window based statistical strategy to extract users’ historical statistical features \cite{ref7, ref8, ref9, ref10, ref11, ref12, ref13, ref14, ref15, ref16, ref17, ref18, ref19, ref20, ref21}. They use a fixed time period as a fixed window to perform statistical operations within the window, such as the number of purchases, the number of views, and the total amount of consumption for a product within a fixed time window like last 2 days, last 3 weeks or last 2 months etc. This method will lose a lot of detailed information, especially the sequence information and the time interval information, just like the bag of words model in the NLP field, it loses the sequence information of words, which makes it difficult to improve the prediction accuracy.
\par
In practice, we observe that users’ historical daily behaviors operated on the mobile banking APP can be used as a supplement of user information. In the mobile banking APP, users will have behaviors such as wealth management, purchasing funds and transferring, which can be helpful to reflect the users’ intentions and habits that contribute to constructing users’ credit profile. For example, users' behaviors in short-term sessions can indicate whether the user is deliberate or impulsive. Especially the last session just before applying a loan, a shorter number of operations and time intervals may indicate that the users’ intention state is more impulsive, thus this will lead to higher overdue rate. And users with high consumption habits who like to buy high-risk and high-yield products have a more aggressive attitude towards finance, thus their overdue rate will be relatively higher; while those with low consumption habits who prefer to buy low-risk and low-yield products have a relatively conservative attitude towards finance, thus their overdue rate is relatively lower. Therefore, the behavior sequences before applying for a loan can reflect the users’ short-term intention within a session and users’ long-term habits, both of which will have an impact on the final overdue rate.
\par
To capture sequence information of user behavior in details and avoid the shortcomings of statistical strategy in traditional models, we propose a model called HUIHEN (Hierarchical User Intention-Habit Extract Network). Considering the diversity of users’ behaviors in the mobile banking APP, we divide user behaviors into sessions to ensure the consistency and continuity of behaviors in the same session and conduct embedding operations on these behaviors. For users' behaviors have different types and each type of operation holds a different meaning, we adopt a field-aware method to extract the intra-field information of the behaviors \cite{ref22}. Then, to effectively capture the users’ behavior information of the short-term intentions and long-term habits, we design a hierarchical architecture composed of time-aware GRU and user-item-aware GRU. Time-aware GRU is used to extract the intention state information, which reflects the users’ short-term intention information within a session, and the last session information can accurately reflect the users’ intention state just before the application. User-item-aware GRU takes the output results of time-aware GRU for each session as input to extract users' long-history information. While in the early period of online service, there are relatively not enough labeled data containing behavior sequences, inspired by transfer learning and teacher-student architecture, we introduce a transfer auxiliary training network to assist the training of the entire model to further improve the accuracy. In above ways, without increasing the complexity of the application process and requiring user to fill in more personal information, our model can achieve better results to predict the overdue rate.
\par
The main contributions of our work can be summarized as follows:
\par
\begin{itemize}
\item We propose a new way that leverages users’ detailed historical behavior sequences in mobile banking APP to extract information for credit loan overdue risk detection for the first time, which can be regarded as a supplement of user credit profile without increasing the complexity of application process.
\item We propose a method to merge and abstract the users’ behavior in the mobile banking APP, and leverage the field-aware method to vectorize them, which reduces the size of model parameters and facilitates model training.
\item We propose a hierarchical architecture composed of time-aware GRU which is used to extract users’ short-term intention within a session and user-item-aware GRU which is used to extract users' long-term habits. Both short-term intentions and long-term habits are meaningful to overdue detection.
\item Experimental results have confirmed that our model can obtain better results and achieve the state of art performances on all datasets.
\end{itemize}
\section{Related Work}
Nowadays, credit risk is the most critical challenge in the banking industry, and credit risk detection has attracted more attention from academic research. To this end, many classification methods have been proposed so far \cite{ref23,ref24,ref25,ref26,ref27}. These methods can be divided into two categories: (1) rule-based methods and (2) model-based methods. The first type of method combines different techniques to propose a rule-based evaluation method \cite{ref28, ref29}. Although rule-based methods are easy to understand and implement, the new type of risk will not be correctly classified after rules are generated \cite{ref30}. The second category includes statistical methods \cite{ref9} and artificial intelligence methods \cite{ref10, ref11}, such as logistic regression, Support Vector Machine \cite{ref31, ref32} and neural networks \cite{ref16, ref18}. Danenas and Garsva \cite{ref33} introduced a new approach based on linear SVM combined with external evaluation and sliding window testing for large data sets. Furthermore, experiments have shown that their methods provide the same excellent results as logistic regression and RBF networks. Geng et al. \cite{ref34} used a neural network, Decision trees, and SVM to build an early warning system that serves to forecast the financial distress of Chinese companies. The obtained results show that the neural network is more accurate than the other classifiers.
\par
The neural networks models with deeper layers can acquire features directly from the training dataset and learn the distribution of the data. Luo et al. \cite{ref35} applied Deep Belief Network (DBN) with Restricted Boltzmann Machines to examine the performances of credit scoring models. When comparing to other traditional techniques such as logistic regression and SVM, the authors found out that deep belief networks gave the best performance. Zhang et al. \cite{ref36} studied the cross border e-commerce credit risk assessment model based on the neural network of particle swarm optimization genetic algorithm. Tavana et al. \cite{ref37} used both Artificial Neural Networks (ANNs) and Bayesian Networks (BNs) methods to assess financial issues related to key factors in liquidity risk. The authors concluded that these two technologies were quite complementary in revealing liquidity risks. Huang et al. \cite{ref38} proposed the probabilistic neural network (PNN) which had the minimum error rate and second type of errors. However, the above model only uses the users’ static statistical characteristics as user credit profile to learn credit risk, which is insufficient to obtain the accurate overdue rates results.
\par
With the continuous development of deep learning, more and more deep learning technologies are used in such as recommendation systems, advertisement click rate estimation, and many natural language processing tasks \cite{ref54, ref55}. Recurrent neural networks (RNN) such as LSTM \cite{ref39} and GRU \cite{ref40} are proposed for processing sequence models. The performance of GRU is comparable to LSTM, but the parameters are less than LSTM, so it can be trained faster and requires less training data. One of the important reasons why deep learning can achieve great success in the field of recommendation system is to make full use of the users’ product behavior, and tap out the users’ real interest shown by the user behavior \cite{ref41, ref42}. In our work, we leverage user historical behaviors in the mobile banking APP to extract users' short-term intentions and long-term habits, which are important features to calculate the users’ overdue risks.
\par
In recent years, teacher-student strategies have gradually become popular. Hinton et al. \cite{ref43} and Romero et al. \cite{ref44} proposed knowledge distillation and FitNet, respectively. Zhou et al. \cite{ref45} proposed a general framework named rocket launching to get an efficient well-performing light model with the help of a cumbersome booster net. They used light networks with fewer parameters and layers to decrease the inference time, which was helped by a pre-trained complex teacher network that trained in advance. Transfer learning has attracted a lot of attention in recent years \cite{ref46, ref47}. The objective of transfer learning is to make these higher level representations more abstract, with their individual features more invariant to most of the variations that are typically present in the training distribution, while collectively preserving as much as possible of the information in the input \cite{ref48}. Inspired by transfer learning and teacher-student architecture, the whole training process proposed in this paper is divided into two steps. Part of parameters of the model are pre-trained in first step in which the teacher-student like architecture is leveraged to further improve performance.
\par
\section{Hierarchical User Intention-Habit Extract Network}
\subsection{The Overall Structure of HUIHEN}
\begin{figure}[h!]
	\centering
	\includegraphics[width=\textwidth]{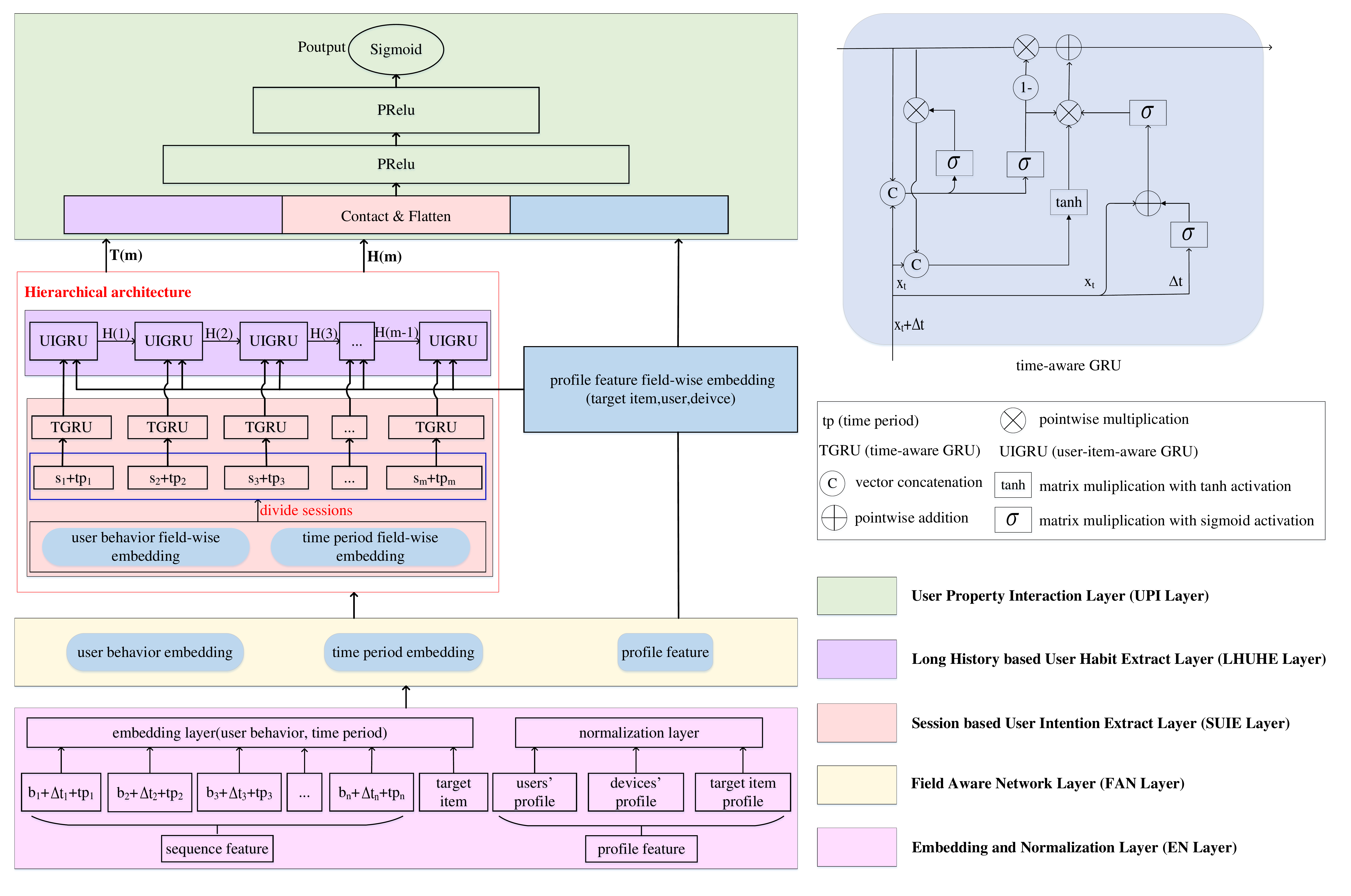}
	\caption{overall architecture of HUIHEN model.}\label{model}
\end{figure}

As shown in Figure \ref{model}, the proposed model HUIHEN includes five parts, which successively complete key tasks of feature vectorization, short-term and long-term behavior features extraction, feature fusion and make the final prediction.
\par
The input layer of model HUIHEN is EN Layer (Embedding and Normalization Layer), which obtains the numerical form of the user's sequence feature and profile feature to facilitate the processing of the latter part. Its input features include variable-length sequence feature and fixed-length profile feature. This layer performs embedding operation on sequence feature and encodes profile feature into the fixed-length vector form. 
\par
To easily extract intra-field features belonging to different categories (categories are divided according to types of sequence feature, detailed division way can be seen in Section \ref{section3.2}, and intra-field features refer to the information contained in users’ behavior category attributes), FAN Layer (Field Aware Network Layer) performs different mapping operations based on the field of outputs from the previous layer and obtains field-wise features. 
\par
To delicately extract the users' short-term intention attributes and long-term habits information from the users' behavior sequences, HUIHEN proposes a hierarchical architecture (the red solid line box in Figure \ref{model}) consisting of SUIE Layer (Session based User Intention Extract Layer) and LHUHE Layer (Long History based User Habit Extract Layer) to extract features from field-wise data of users’ sequence feature in different time windows. 
\par
SUIE Layer adopts time-aware GRU to extract users’ short-term intentions combined with time period and time interval information in each session. LHUHE Layer adopts user-item-aware GRU to extract users’ long-term habits combined with user-item information $H(m)$ based on the intention sequence. 
\par
Finally, to fully consider user information in prediction of overdue rate, UPI Layer (User Property Interaction Layer) fuses sequence information extracted in the hierarchical architecture together with profile features. It takes extracted users’ short-term intention in last session $T(m)$, long-term habits $H(m)$ and other profile feature as input, then extracts the interaction relationship between all user features through MLP (Multi-Layer Perceptron) network.
\subsection{Embedding and Normalization Layer}\label{section3.2}
To maximally retain the original user sequence information, this layer utilizes embedding method to vectorize users’ behavior sequences. It can be seen that users can perform diverse types of business operations on mobile banking APP, such as transferring, purchasing fund, and consumption. Different business operations generate different behavior sequence, thus types of user behaviors contained in mobile banking APP are diverse. If performing embedding operation directly on original behaviors, it will add redundant information towards diverse behaviors, and due to the influence of long-tails distribution, the parameters of model will increase (behavior embeddings belong to the part of the model’s parameter), which is not conducive to train the final model. To make the model focus on information of behaviors that can reflect the users' credit attributes and reduce the size of the model parameters, we abstract and merge user behaviors instead of directly performing embedding operation.
\begin{figure}[h!]
	\centering
	\includegraphics[scale=0.5]{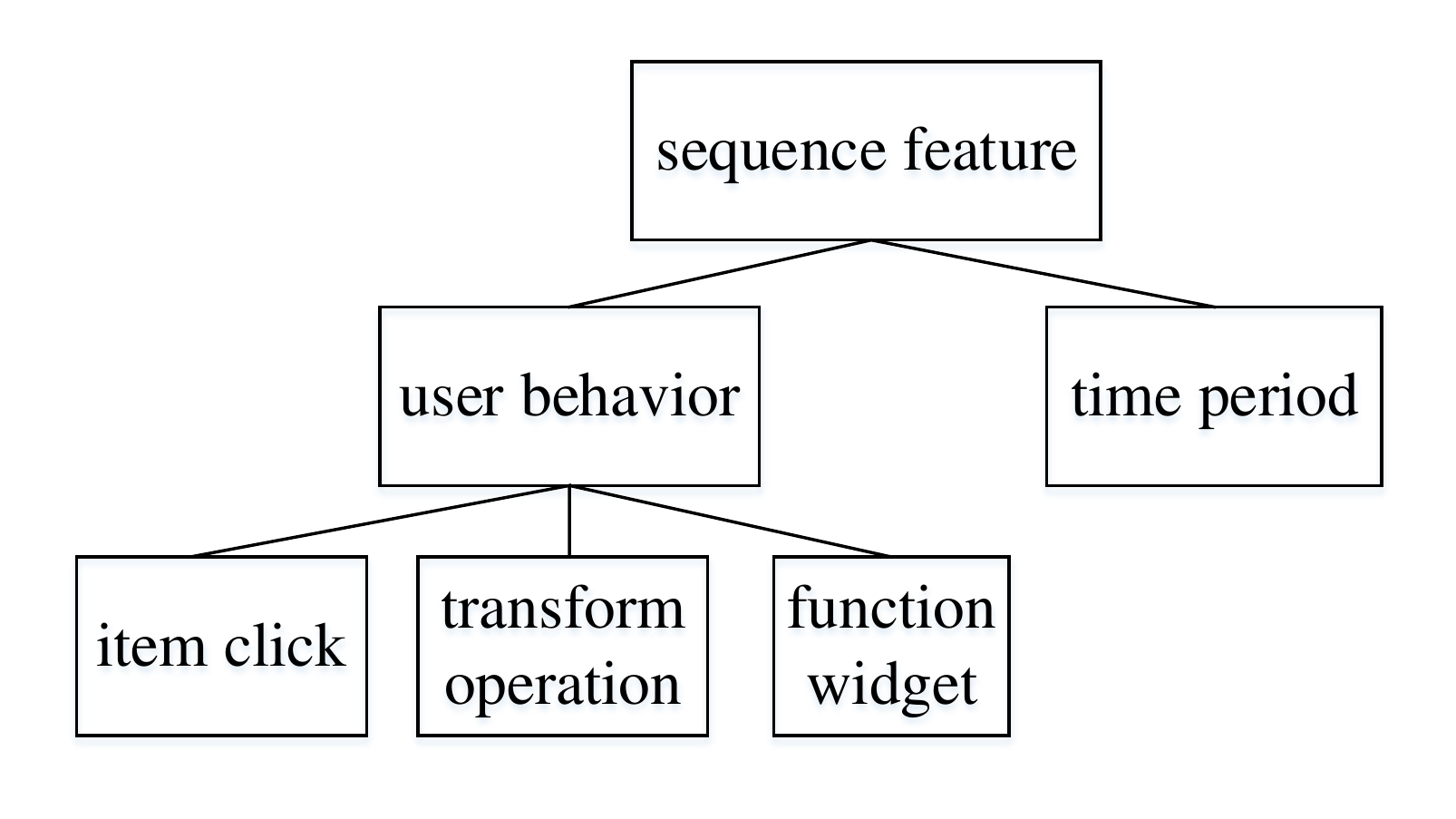}
	\caption{categories of sequence feature.}\label{categories}
\end{figure}
\par
As shown in Figure \ref{categories}, by abstracting and merging based on the attribute of user behaviors, sequence feature is divided into four categories (leaf nodes of the tree structure in Figure \ref{categories}): item click, transform operation, function widget and time period. Item click, transform operation and function widget belong to user behavior ($[{b_1},{b_2},...,{b_n}]$) which reflects users’ behavior information. As shown in Figure1, the time interval between adjacent behaviors $\Delta{t_i}(i=1,2,...,n-1)$ is also included in sequence feature, and is inserted into user behaviors sequence ($[{b_1},\Delta{t_1},{b_2},\Delta{t_2},...,\Delta{t_{n-1}},{b_n}]$). Time period ($[{{tp}_1},{{tp}_2},...,{{tp}_n}]$) reflects the time period, including two types of week and month, which record the time period information when the behavior happens. Item click and transform operation refer to operation behaviors related to financial business in mobile banking APP, where item click record information corresponding to behaviors of clicking and browsing different products, and transform operation mainly includes various amount-related behaviors which represents successfully complete a financial-related business such as purchasing behavior and transferring behavior. To describe users'  habits towards financial products in detail, item click is further divided into credit loan item click, deposit item click, fund item click and so on based on the type of product. Function widget refers to behaviors of clicking function widgets which are not related to business in mobile banking APP. For example, the confirm and return function buttons exist on many pages in the mobile banking APP, but they are not related to financial business. This division way can highlight the main problem to be solved, describe the behavior related to target task in detail, which merge and simplify the behavior that are not related to the target task. Then different embedding methods are adopted on these categories before fed into the following model.
\par
(1) Performing embedding operation on item click obtains item click embedding. Each type of item click embedding corresponds to a parameter matrix ${R^{N*{d_{item}}}}$, where $N$ represents the number of different items, ${d_{item}}$ represents the dimension of item click embedding. (2) Performing embedding operation on transform operation obtains transform operation embedding. Every type of transform operation embedding is represented by expression ($embedding * log(value)$), where value represents the amount of product corresponding to operation. Each type of transform operation may correspond to many different items (such as different type of fund products). But if these items belong to same type (such as all belong to the fund products), they are abstracted and merged into the same one embedding. Taking purchasing financial products as an example, the vector $E$ represents the behavior of purchasing financial product which calls financial transform. When a user buys financial product $A$, and its amount is $amount(A)$, then the final behavior expression is $E * log(amount(A))$. (3) Performing embedding operation on function widget obtains function widget embedding. As mentioned before, the function widget like confirm and return function buttons exist on many pages in the mobile banking APP, but they are not related to financial business. Thus, the function buttons (such as confirm button and return button) on all pages only correspond to one embedding representation. For time period, each date (Monday to Sunday, 1st to 31th) is assigned a corresponding embedding vector. 
\par
Profile feature mainly includes users’ profile, devices’ profile, and target item profile. Each dimension of profile data is unique and certain, so we do not need to abstract and merge them. The first step is to concatenate all the profile features into a vector , the second step is to encode the vector into the numerical form and normalize it using mean and var. By adopting the normalization operation, the distribution of normalized data is unbiased and low variance. 
\par
After being processed by the EN Layer, the original user behaviors $[{b_1},{b_2},$ $...,{b_n}]$ is transformed into a vector sequence $[{e_1},{e_2},...{e_n}]$, while time interval between adjacent behaviors $\Delta{t_i}(i=1,2,...,n-1)$ is also inserted into the vector sequence ($[{e_1},\Delta{t_1},{e_2},\Delta{t_2},...,\Delta{t_{n-1}},{e_n}]$). Overall, the output of this layer includes user behavior embedding (item click embedding, transform operation embedding, function widget embedding), time period embedding, and normalized profile feature.
\subsection{Field Aware Network Layer}\label{section3.3}
Sequence feature has been divided into different categories as shown in Figure \ref{categories}, and each category corresponds to a filed. For sequence feature, since different categories contain important information in their respective fields, FAN Layer captures the intra-field information using those features inside the same field. For profile feature, to facilitate training the model, this layer maps them to the same feature space as users' behavior field-wise feature.
\par
This layer performs different mapping operation on different categories of embedding representation from the previous layer. It introduces a transformation matrix ${w_i}$ for each field, and the behavior embedding ${e_j}$ belonging to a certain field $i$ can obtain field-wise feature as shown in Equation(1). The user behavior embedding sequence $[{e_1},{e_2},...,{e_n}]$ is transformed into the vector sequence $[{f_1},{f_2},...,{f_n}]$. Similarly, the time interval between adjacent behaviors $\Delta{t_i}(i=1,2,...,n-1)$ is inserted into the vector sequence $[{f_1},{f_2},...,{f_n}]$ ($[{f_1},\Delta{t_1},{f_2},\Delta{t_2},...,\Delta{t_{n-1}},{f_n}]$).The time period embedding is mapped into the time period field. For normalized profile feature, this layer adopts the same mapping method (Equation(1)) as user behavior embedding to perform a matrix transformation on the vector. 
\begin{equation}
{FieldAware_i}(e_j) = Relu({w_i}*{e_j})
\end{equation}
\par
Overall, this layer obtains user behavior field-wise embedding (item click field-wise embedding, transform operation field-wise embedding, function widget field-wise embedding), time period field-wise embedding, and profile feature field-wise embedding.
\subsection{Session based User Intention Extract Layer}
The intention of users determines user’s behavior, which is relatively stable within a short period of time (like one hour). The business in mobile banking APP is diverse, thus user intention in different sessions (a short period of time) is different, and user’s behavior pattern contained in different sessions is also inconsistent. Therefore, to extract users' short-term intentions, we first divide the behavior sequence into sessions so that consistent information is contained in a session, and then extract user’s intentions from user behaviors of each session.
\par
SUIE Layer first divides the vector sequence $[{f_1},{f_2},...,{f_n}]$ generated by the FAN Layer into sessions based on the time interval (one hour) between sequence elements. Section \ref{section3.3} has mentioned that the time interval between adjacent behaviors $\Delta{t_i}(i=1,2,...,n-1)$ would be inserted into the vector sequence $[{f_1},{f_2},...,{f_n}]$ and obtain vector sequence like ($[{f_1},\Delta{t_1},{f_2},\Delta{t_2},...,\Delta{t_{n-1}},$ ${f_n}]$). Therefore, this layer can easily divide the vector sequence $[{f_1},{f_2},...,{f_n}]$ into sessions by time interval. After dividing sessions, the original vector sequence $[{f_1},{f_2},...,{f_n}]$ becomes the session sequence $[{s_1},{s_2},...,{s_m}]$, in which each session corresponds to a series of original user behaviors, namely, ${s_i} = [{f_1},\Delta{t_1},{f_2},\Delta{t_2},...,\Delta{t_{k-1}},{f_k}]$. The length of the vector sequence in each session is different, and each session represents a complete user’s operation behavior using the mobile banking APP. 
\par
Then session sequence $[{s_1},{s_2},...,{s_m}]$ is fed into time-aware GRU to independently model behaviors within each session. To obtain more meaningful information by modeling behaviors in each session, the time-aware GRU adds the time information which contains two aspects: the time interval between user behaviors, and the time period information ($x_{week}$ and $x_{month}$) of user’s operations. By considering the time period of user’s behavior, the layer can extract user’s periodic behavior information. For example, people who use up their salary every month are more inclined to apply for credit at the end of the month. In addition, the time interval reflects user’s psychological state and proficiency during the credit application process. For example, if user’s time interval of all behaviors in one session is generally too long, it means that the user is not familiar with the business operations. However, if user’s most time interval of behaviors is relatively short, and a certain behavior has a long-time interval, it means that the user is hesitate facing those operation behaviors. The proposed HUIHEN can more accurately model user behaviors by adding the time interval between behaviors and the time period information. So the time interval ${\Delta}t$ between user’s behaviors, time period vector $[{x_{week}};{x_{month}}]$, and the session sequence $[{s_1},{s_2},...,{s_m}]$ are fed into the time-aware GRU to model user’s intention in each session. Formulas of the time-aware GRU are shown in Equation(2) to Equation(6) below:
\begin{gather}
{u_t} = \sigma ({w_u}\cdot \left[{x_t};{h_{t-1}};{x_{week}};{x_{month}}\right] + {b_u})\\
{r_t} = \sigma ({w_r}\cdot \left[{x_t};{h_{t-1}};{x_{week}};{x_{month}}\right] + {b_r})\\
{p_t} = tanh({w_p}\cdot \left[{x_t};{h_{t-1}};{x_{week}};{x_{month}};{r_t} \odot {h_{t-1}}\right] + {b_p})\\
{T_m} = \sigma ({w_m}\cdot[{x_t} + \sigma ({Q_t}{\Delta {t_m}})] + {b_m})\\
{h_t} = (1-{u_t})\odot{h_{t-1}} + {p_t}\odot{T_m}\odot{u_t}
\end{gather}
Where the unit of $\Delta{t_m}$ in Equation(5) is second, the value of time may be relatively larger. To shrink the output range, the sigmoid activation function is added to the Equation(5) and the $\Delta{t_m}$ is changed to $\sigma ({Q_t}{\Delta{t_m}})$. Compared with the traditional GRU, the time-aware GRU adds the new input information and the new gated unit. According to Equation(2) and Equation(3), the time-aware GRU adds time period vector $[{x_{week}};{x_{month}}]$ in the input part, so that the time period information can affect output results of the update gate and reset gate. In addition, the time-aware GRU constructs a gated unit ${T_m}$ that can reflect the time interval as shown in Equation(5), thus the time interval between user behaviors can control the size of information input to the next moment.
\par
Abstracting Equations ((2) to (6)), each session ${s_i}$ obtains the output  as shown in Equation(7).
\begin{align}
{T_i} = time-aware~GRU({s_i}) = & time-aware~GRU({p_k}~in~{s_i}, week, month,\Delta t)\notag \\
& i = 1,2,...,m
\end{align}
\par
Finally, the entire session sequence $[{s_1},{s_2},...,{s_m}]$ is transformed into $[{T_1},{T_2},$ $...,{T_m}]$, where ${T_i}\in {R^{d_{TGRU}}}$ is a vector, and its dimension is consistent with the output dimension of the time-aware GRU hidden layer.
\subsection{Long History based User Habit Extract Layer}
Each element $T_i$ in the output sequence $[{T_1},{T_2},...,{T_m}]$ of the time-aware GRU represents the user’s intention information in the corresponding session, and $[{T_1},{T_2},...,{T_m}]$ constitutes an intention sequence. Based on the fact that a series of user intention behaviors reflect user habits, this layer adopts user-item-aware GRU to model these intention sequences to learn users' long-term habits towards target item. In addition, the user-item-aware GRU can extract the change and evolution of user-related attributes from the context of elements in intention sequence, including changes in objective factors (personal financial situation) and subjective factors (personal attitude towards financial affairs), which is conducive to more accurately describe user. Equations of user-item-aware GRU are shown in (8) to (11) below:
\begin{gather}
{u_t} = \sigma ({w_u}\cdot \left[{x_t};{h_{t-1}};{x_{user}};{x_{item}}\right] + {b_u})\\
{r_t} = \sigma ({w_r}\cdot \left[{x_t};{h_{t-1}};{x_{user}};{x_{item}}\right] + {b_r})\\
{p_t} = tanh({w_p}\cdot \left[{x_t};{x_{user}};{x_{item}};{r_t} \odot {h_{t-1}}\right] + {b_p})\\
{h_t} = (1-{u_t})\odot{h_{t-1}} + {p_t}\odot{u_t}
\end{gather}
\par
Compared with the traditional GRU, user-item-aware GRU adds $x_{user}$ and $x_{item}$ information in the input part, which represents the attribute information of users and the applied credit product item. For different users and different applied credit products, each element in intention sequence has a different contribution to the prediction of overdue rate. By adding information of these two parts, the model proposed in this paper can purposefully extract information about the current user and the applied credit product from intention sequence. Abstracting Equations ((8)-(11)), the final output of this layer is shown in Equation(12):
\begin{equation}
{H_m} = user-item-aware~GRU(T,user,item)
\end{equation}
\par
The output $H_m$ of user-item-aware GRU at the last moment represents the users' long-term habits extracted from the users' operation behavior in the past long period, and describes the users' attitude and character towards financial products.
\subsection{User Property Interaction Layer}\label{upi}
The proposed hierarchical architecture captures the user's short-term intention within a session and users' long-term habits. To fully consider user information in prediction of overdue rate, UPI Layer fuses the features extracted in the hierarchical architecture together with profile features, which aims to mine the interaction relationship between profile feature and sequence feature. Thus, the user’s intention state just before applying for a loan $T(m)$ extracted from time-aware GRU and users' long-term habits $H(m)$ extracted from user-item-aware GRU are concatenated together with profile features such as user portraits and used as the input of the subsequent full connected network. The activation function PRelu as shown in Equation(14) is used in the full connected network. Finally, this layer obtains the final output $P_{output}$ as shown in Equation(13) that represent the overdue probability. The loss function of model HUIHEN is shown in Equation(15).
\begin{gather}
{P_{output}} = sigmoid(PRELU(MLP(x)))  \\
PRELU(x) = max(0,x) + \alpha * min(0,x) \\
\begin{align}
{L_{cross\_entropy}}({P_{output}}(x)) = & -({y_{lable}}log({P_{output}}(x))~+ \notag \\
& (1 - {y_{label}})log(1 - {P_{output}}(x)))
\end{align}
\end{gather}
Where $\alpha$ in Equation(14) is a hyper-parameter with an initial value of 0.5, which will also be trained during training phase. ${P_{output}}(x)$ in Equation(15) represents the overdue prediction probability of the proposed model for sample $x, {y_{label}} \in {0,1}$ represents the true label for sample $x$.
\section{Training Strategy}\label{section4}
\subsection{Background Analysis}
As the previous overdue risk detection system only collected and leveraged the profile features, the mobile banking APP needs to be updated to extra collect user behavior sequence information. In this paper, the data before the APP updated is called old data, and the data after the APP updated is called new data. Old data only contains profile features while new data contains both sequence features and profile features. In the short time after APP is updated, there are actually not enough labeled samples with user behavior sequences, and a small number of samples are not conducive to the training of the entire model. Since many labeled samples without user behavior sequences have been accumulated before the update, these samples can be used to pre-train the parameters of the UPI Layer. Referring to the idea of transfer learning \cite{ref46,ref47,ref48}, the training process is divided into two steps. In first step, the network parameters of the UPI Layer are pre-trained using the old data by a transfer auxiliary training network. In addition, inspired by the idea of teacher-student architecture \cite{ref43, ref49, ref50}, a teacher-student like architecture is used to further improve the performance of the pre-trained model in the transfer auxiliary training network which will be discussed in Section \ref{section4.2} in detail. In second step, the new data is used to train the entire model based on the pre-trained model parameters of the UPI Layer.
\subsection{Transfer Auxiliary Training Network for Pre-training}\label{section4.2}
\begin{figure}[h!]
	\centering
	\includegraphics[scale=0.3]{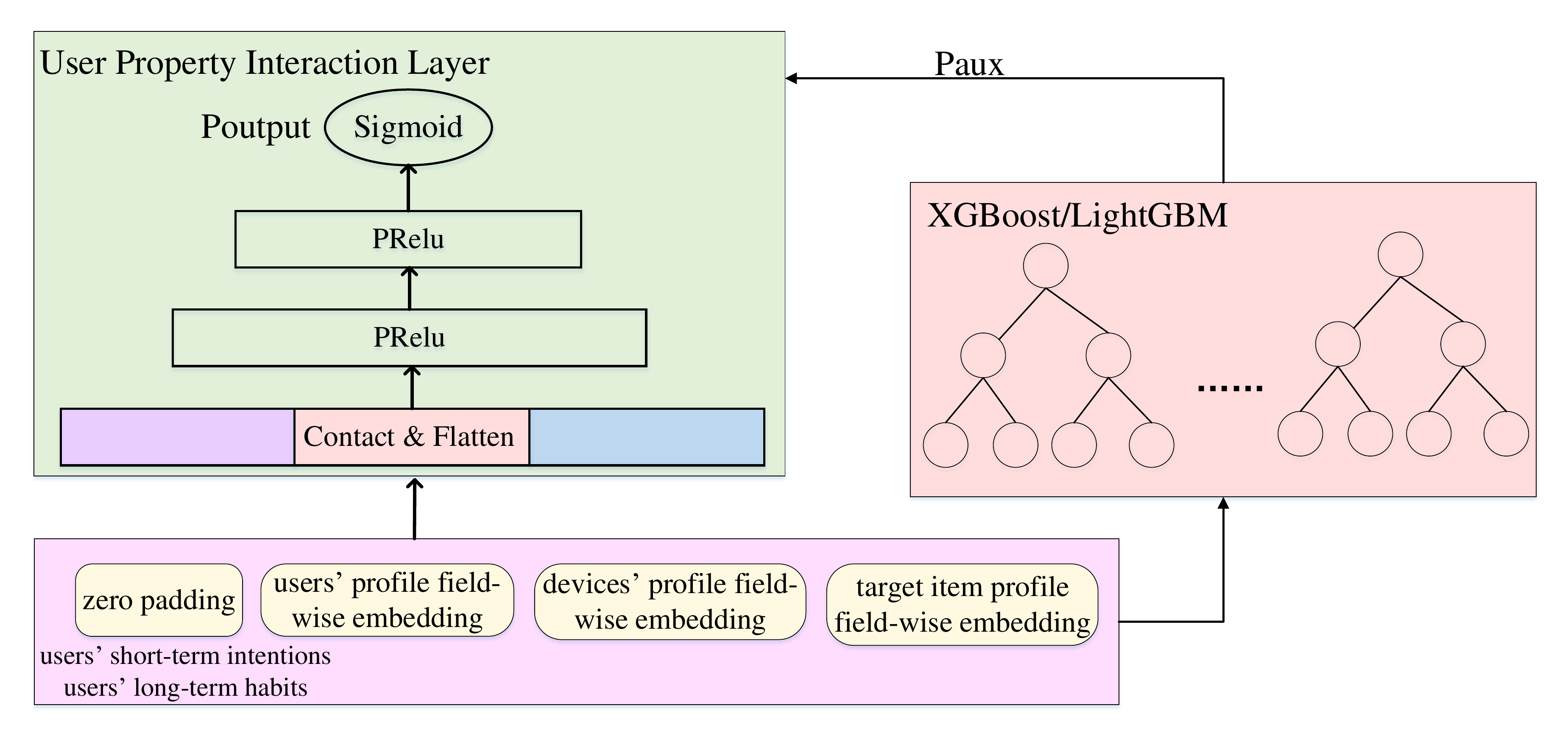}
	\caption{the structure of the transfer auxiliary training network.}\label{auxilary}
\end{figure}
\par
According to the structure of the transfer training auxiliary network shown in Figure \ref{auxilary}, this section makes a careful description of the first step in training process. The sequence features from the time-aware GRU and user-item-aware GRU are filled with zero vector of the same dimension to keep the feature dimensions fed into the following UPI Layer when using old data that does not contain user behavior sequences. In this way, when the neural network propagates forward, the value of this part will not have any effect on the output of these neurons (including positive and negative effects). During pretraining, the UPI Layer can learn the relationship between profile features and the overdue rate, and when new data is used for subsequent model training, the entire model can pay more attention to the influence of user behavior sequence information, which is more helpful to model training. 
\par
For structured features generated by artificial feature engineering, the tree-based model Xgboost/LightGBM has a better prediction performance than MLP\footnotemark \footnotetext{\fontsize{5pt}{0.5em}\selectfont For structured features generated by artificial feature engineering, tree based model as Xgboost/LightGBM have a better performance than MLP network. However, for unstructured data such as click behavior sequence, deep learning can automatically learn the more detailed feature such as information within jumps in click behavior sequence. In addition, if features are only generated by artificial feature engineering, the more detailed features from unstructured data will be lost, namely, the unstructured data is crucial. Therefore, in the first step, Xgboost/LightGBM can be used to generate the guidance probability when training model with old structured dataset; in the second step, the new dataset which is used to train the model includes lots of unstructured data, it is better to adopt hierarchical-based custom GRU which belongs to one of the deep learning architectures.\par}, so Xgboost/LightGBM can be used as a teacher of MLP for guidance. Thus, Xgboost/LightGBM are firstly trained using old data, and then Xgboost/LightGBM will generates a guidance probability for each sample $x$. The guidance probability can be called as the soft label which usually contains more information than the original hard label (0/1). For example, when the prediction probability of sample A and sample B are o.6 and 0.9 respectively, although their prediction labels are all 1, their degree of bias to 1 is different, that is, the information provided to the model to learn is also different. The loss function of MLP containing label information and guidance probability information is shown in Equation(16).
\begin{align}
{Loss_{aux}}(x)~=~& {L_{cross\_entropy}}({P_{output}}(x)) + \frac{\alpha}{2} (KL({P_{output}}(x), {P_{aux}}(x))~+ \notag \\
& KL({P_{aux}}(x), {P_{output}}(x)))
\end{align}  
Where $\alpha$ in Equation(16) is a hyper-parameter with an initial value of 1, which will also be trained during training phase.
\section{Experiments}
\subsection{Dataset}
We collect two datasets (${Data}_A$ and ${Data}_B$) from Bank A and Bank B\footnotemark \footnotetext{\fontsize{5pt}{0.5em}\selectfont Because it involves user privacy information, the bank name is replaced by Bank A and Bank B. \par} in China, including the old dataset and the new dataset. The old dataset only contains profile feature (such as user profile, device profile) before updating the mobile banking APP, and the new dataset adds sequence feature (such as purchase behavior, click behavior) after updating the mobile banking APP.
\par
It is clear that before updating the SDK in the mobile banking APP, the original online overdue risk detection system of bank just depend on these profile features. ${Data}_{oldA}$ represents the old dataset of Bank A before updating the mobile banking APP, and ${Data}_{newA}$ represents the new dataset of Bank A after updating the mobile banking APP. ${Data}_{oldB}$ represents the old dataset of Bank B before updating the mobile banking APP, and ${Data}_{newB}$ represents the new dataset of Bank B after updating the mobile banking APP. ${Data}_A$ includes ${Data}_{oldA}$ and ${Data}_{newA}$. ${Data}_B$ includes ${Data}_{oldB}$ and ${Data}_{newB}$. Table \ref{dataset-table} describes the sample number of datasets, where Negative represents the sample number of overdue data and Positive represents the sample number of normal data, and Total indicates the total sample number of dataset.
\begin{table}[htbp]
	\centering
	\caption{the number of samples of related datasets}\label{dataset-table}
	\begin{tabular}{cccc}
		\hline
		dataset & Negative & Positive & Total\\
		\hline
		${{Data}_{newA}}$ & 1402 & 25200 & 26602\\
		${{Data}_{oldA}}$ & 9501 & 133020 & 142521\\
		${{Data}_{newB}}$ & 734 & 12031 & 12765\\
		${{Data}_{oldB}}$ & 982 & 13290 & 14272\\
		\hline
	\end{tabular}
\end{table}
\par
\subsection{Experiments Setup}
Before evaluating the proposed model, we present experiment setup information in this section. As mentioned in Section \ref{section3.2}, user behaviors need to be abstracted and merged. Taking transform operation and function widget as example, if we perform embedding operation directly on each behavior, roughly 300-400 behavior embedding representations will be generated. Therefore, this paper abstracts and merges these behaviors based on its attribute, and finally the model HUIHEN obtains about 40-50 different number of embeddings.
\par
The time interval for session extraction is one hour, that is, when the time interval between two behaviors exceeds one hour, they are divided into two different sessions. Because the number of sessions is different, this paper chooses the 50 most recent effective sessions (sessions with less than 3 behavior sequences are filtered as noise) to represent user behavior information, and 50 sessions basically cover 3-6 months of user behaviors. The proposed model in this paper will retain the latest 50 sessions for users with more than 50 sessions, and fill session sequence for user with less than 50 sessions using specific identification vector such as zero vector. To facilitate training model, our proposed model only keeps the latest 25 behaviors in each session before the session is fed into the model, and will fill the head of those session with less than 25 behaviors using specific identification vector such as zero vector. The aforementioned number of user sessions and the number of behaviors in each session are all obtained by using 5-fold cross validation. 
\par
The dimension of user behavior embedding is 10, thus the dimension of transformation matrix $w_i$ in Section \ref{section3.3} is $R^{10*10}$. The output dimension of user-item-aware GRU and time-aware GRU are both set to 30. UPI Layer includes three fully connected layers, and the neurons are set to 50, 25 and 2, respectively. The Dropout whose parameter is set to 0.6 is adopted to mitigate the overfitting problem. The proposed model is implemented by tensorflow \cite{ref51}, and optimizer is Adam \cite{ref52} whose learning rate is 0.001. AUC and KS are used as the evaluation standard, and KS is the evaluation standard often used in the credit field which reflects the ability to distinguish positive and negative samples.
\subsection{Experiment Result and Evaluation}
\subsubsection{Performance Comparison of Proposed Model with Other Models}\label{sy}
In order to prove the effectiveness of our proposed model, we construct a complete compared model set which consists of some variant models derived from standard HUIHEN and some traditional machine learning models based on profile features.  The compared model set include HUIHEN-withoutTUGRU-AuxNet-xgb, HUIHEN-withoutAuxNet, HUIHEN-AuxNet-withoutxgb, GRU-AuxNet-xgb, Xgboost, LR, MLP, RandomForest and LightGBM. HUIHEN-withoutAuxNet represents the model that has not been pre-trained with transfer training auxiliary network. HUIHEN-AuxNet-withoutxgb adopts the transfer training auxiliary network, but the loss function directly uses the original label and does not include the guidance probability generated by Xgboost/Lightgbm. GRU-AuxNet-xgb does not divide behavior sequence into session based on time interval of behaviors, but directly feed user behavior sequence into GRU, which can be regarded as a normal deep learning architecture to model user behavior sequence. HUIHEN-withoutTUGRU-AuxNet-xgb replaces time-aware GRU and user-item-aware GRU in HUIHEN with traditional GRU, that is, it does not consider user-item information and time period during model training. The input features of Xgboost, LR, MLP, RandomForest and LightGBM are similar to that of the transfer training auxiliary network in Section \ref{section4}. HUIHEN-withoutAuxNet is trained using the new dataset because it is not pre-trained with transfer training auxiliary network. Xgboost, LR, MLP, RandomForest and LightGBM can’t accept unstructured input like user behavior sequence, thus user behavior information of the new dataset is removed when training these models. Table \ref{table2} and Table \ref{table3} show the comparison results of the proposed model in this paper and these models on ${Data}_A$ and ${Data}_B$.

According to the above experimental results, the conclusions are as follows:
\par
(1) From Table \ref{table2} and Table \ref{table3}, we can obviously see that the proposed model HUIHEN outperforms the HUIHEN-withoutAuxNet. By using HUIHEN, AUC and KS on ${Data}_A$ increase by 1.41\% and 2.74\%, and AUC and KS on ${Data}_B$ increase by 0.77\% and 1.64\%, respectively, which shows that the pre-training step of transfer training auxiliary network is effective.
\par
(2) It can be found that the detection performance improvement on ${Data}_A$ is more obvious than on ${Data}_B$. ${Data}_{oldB}$ contains 14,272 samples, while ${Data}_{oldA}$ contains 142,521 samples. Therefore, the sample number of the pre-training step affects the final detection performance. That is to say, the larger sample number in pre-training step, the better the final detection performance.
\par
\begin{table}[htbp]
	\centering
	\caption{Performance Comparison of Proposed Model and Other Models on ${Data}_A$}\label{table2}
	\begin{tabular}{cccc}
		\hline
		Model & AUC & KS\\
		\hline
		HUIHEN & 0.6887 & 0.2845\\
		HUIHEN-withoutTUGRU-AuxNet-xgb & 0.6832 & 0.2795\\
		HUIHEN-withoutAuxNet & 0.6791 & 0.2769\\
		HUIHEN-AuxNet-withoutxgb & 0.6812 & 0.2776\\
       GRU-AuxNet-xgb & 0.6801 & 0.2772\\
       Xgboost & 0.6743 & 0.2645\\
       LR & 0.6245 & 0.2212\\
       MLP & 0.6614 & 0.2589\\
       RandomForest & 0.6691 & 0.2604\\
       LightGBM & 0.6749 & 0.2648\\
		\hline
	\end{tabular}
\end{table}
\par
\begin{table}[htbp]
	\centering
	\caption{Performance Comparison of Proposed Model and Other Models on ${Data}_B$}\label{table3}
	\begin{tabular}{cccc}
		\hline
		Model & AUC & KS\\
		\hline
		HUIHEN & 0.6632 & 0.2543\\
		HUIHEN-withoutTUGRU-AuxNet-xgb & 0.6626 & 0.2537\\
		HUIHEN-withoutAuxNet & 0.6581 & 0.2502\\
		HUIHEN-AuxNet-withoutxgb & 0.6593 & 0.2509\\
       GRU-AuxNet-xgb & 0.6597 & 0.2512\\
       Xgboost & 0.6524 & 0.2486\\
       LR & 0.6213 & 0.2241\\
       MLP & 0.6489 & 0.2394\\
       RandomForest & 0.6501 & 0.2396\\
       LightGBM & 0.6523 & 0.2485\\
		\hline
	\end{tabular}
\end{table}
\par
(3) HUIHEN obtains the AUC of 0.6887 on ${Data}_A$, which is 1.10\% higher than using HUIHEN-AuxNet-withoutXgb (0.6812). HUIHEN obtains the AUC of 0.6632 on ${Data}_B$, which is 0.59\% higher than using HUIHEN-AuxNet-withoutXgb (0.6593). HUIHEN obtains the KS of 0.2845 and 0.2543 on ${Data}_A$ and ${Data}_B$, respectively. Compared with HUIHEN, the KS of HUIHEN-AuxNet-withoutXgb decreases by 1.36\% and 2.49\% on ${Data}_B$ and ${Data}_A$, respectively. Therefore, the detection performance of HUIHEN is better than that of HUIHEN-AuxNet-withoutXgb on two datasets, indicating that the guidance probability generated by teacher-student like training architecture in the pre-training step is effective. Later we will conduct more experiments to further prove the universality of this teacher-student like training architecture.
\par
(4) HUIHEN obtains the AUC of 0.6887 on ${Data}_A$, which is 1.26\% higher than using GRU-AuxNet-xgb (0.6801). HUIHEN obtains the AUC of 0.6632 on ${Data}_B$, which is 0.53\% higher than using GRU-AuxNet-xgb (0.6597). GRU-AuxNet-xgb obtains the KS of 0.2772 and 0.2512 on ${Data}_A$ and ${Data}_B$, respectively. Compared with GRU-AuxNet-xgb, the KS of HUIHEN increases by 2.63\% and 1.23\% on ${Data}_A$ and ${Data}_B$, respectively. Therefore, HUIHEN obtains the better detection performance on two datasets, indicating that it is necessary to divide user behavior sequence into session and adopt hierarchical architecture.
\par
(5) Compared with HUIHEN, the KS of HUIHEN-withoutTUGRU-AuxNet-xgb decreases by 1.79\% on ${Data}_A$ and decreases by 0.24\% on ${Data}_B$. In addition, the model HUIHEN also obtains the higher AUC than HUIHEN-withoutTUGRU-AuxNet-xgb on ${Data}_A$ and ${Data}_B$. Therefore, the detection performance of HUIHEN is better than that of HUIHEN-withoutTUGRU-AuxNet-xgb, which shows that adding user-item information and time period information in GRU can help model accurately identify user behavior features.
\par
(6) Xgboost, LR, MLP, RandomForest and LightGBM obtain the AUC of 0.6743, 0.6245, 0.6614, 0.6691 and 0.6749 on ${Data}_A$, respectively. Compared with HUIHEN, the AUC of these models decreases by 2.14\%, 10.28\%, 4.13\%, 2.93\% and 2.04\%, respectively. Xgboost, LR, MLP, RandomForest and LightGBM obtain the AUC of 0.6524, 0.6213, 0.6489, 0.6501 and 0.6523 on ${Data}_B$, respectively. Compared with HUIHEN, the AUC of these models decreases by 1.66\%, 6.74\%, 2.2\%, 2.02\% and 1.67\%. Similarly, the model HUIHEN obtains the higher KS than  Xgboost, LR, MLP, RandomForest and LightGBM on ${Data}_A$ and ${Data}_B$. Therefore, the detection performance of HUIHEN is obviously better than that of these traditional machine learning models combined with some statistical operation features. This fully shows that the historical behavior of users using the mobile banking APP contains rich information and reflects the user's behavior habits, and this is very meaningful for judging the overdue rate of users applying for credit products. In addition, the detection performance of LightGBM/Xgboost is better than that of LR/MLP/RandomForest on these two datasets, thus this further proves the rationality of using Xgboost/lightGBM to generate the guidance probability in the transfer training auxiliary network.
\subsubsection{Effectiveness of Abstracting and Merging Operations based on Business}
The mobile banking APPs of Bank A and Bank B contain 452 and 367 types of original business operation behaviors (transform operation and function widget), respectively. The operation behaviors of Bank A and Bank B reduce to 42 and 37 business operation behaviors after abstracting and merging, respectively. Table4 shows the detection performance of HUIHEN using original operation behaviors and post-processing operation behaviors.
\begin{table}[htbp]
	\centering
	\caption{The Detection Performance of HUIHEN Using Original Business Operations and Post-Processing Business Operations}\label{table4}
	\begin{tabular}{cccc}
		\hline
		dataset & Operation num & AUC & KS\\
		\hline
		${Data}_A$ & 42 & 0.6887 & 0.2845\\
		${Data}_A$ & 452 & 0.6832 & 0.2786\\
		${Data}_B$ & 37 & 0.6632 & 0.2543\\
		${Data}_B$ & 367 & 0.6582 & 0.2502\\
		\hline
	\end{tabular}
\end{table}
\par
According to Table \ref{table4}, HUIHEN obtains the AUC of 0.6887 using 42 operation behaviors for Bank A, which is 0.81\% higher than using 452 original operation behaviors (0.6832). HUIHEN obtains the AUC of 0.6632 using 37 operation behaviors for Bank B, which is 0.76\% higher than using 367 operation behaviors (0.6582). In addition, HUIHEN obtains the KS of 0.2845 using 42 operation behaviors on ${Data}_A$, and obtains the KS of 0.2543 using 37 operation behaviors on ${Data}_B$. Compared with the HUIHEN using processed operation behaviors, the KS of HUIHEN using original operation behaviors (0.2786) decreases by 2.12\% on ${Data}_A$, and the KS of HUIHEN using original operation behaviors (0.2502) decreases by 1.64\% on ${Data}_B$. Therefore, the HUIHEN using the processed operation obtains better detection performance, which shows that the original business operation behaviors contain more noise and these operation behaviors are reluctant. In addition, if embedding is pesrformed for each operation, it will lead to high sparseness of data, and introduce more training parameters to the model, which will affect the performance of the final model. Finally, it is necessary to abstract and merge the original operation behaviors.
\subsubsection{The Influence of The Number of Sessions and The Number of User Behaviors in The Same Session on The Model Performance}
\begin{figure}[h!]
	\centering
	\includegraphics[scale=0.5]{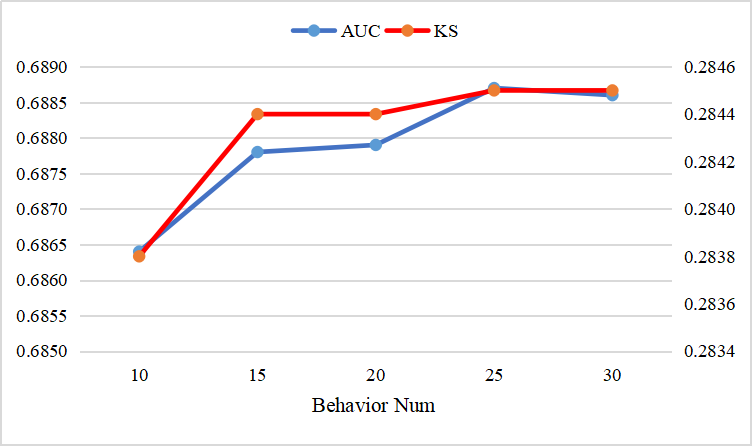}
	\caption{the detection performance of HUIHEN whose number of sessions is set to 50 using different number of user behaviors in the same session.}\label{behavior}
\end{figure}

\begin{figure}[h!]
	\centering
	\includegraphics[scale=0.5]{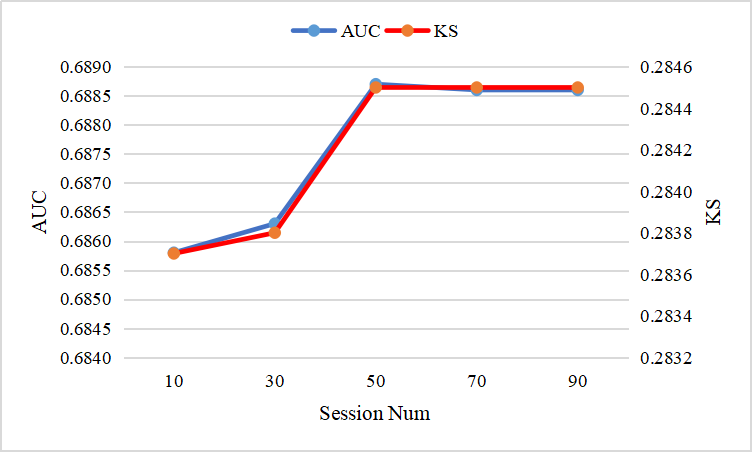}
	\caption{the detection performance of HUIHEN that whose number of user behaviors in the same session is set to 25 using different number of sessions.}\label{session}
\end{figure}

The influence of the number of sessions and the number of user behaviors in the same session on model performance is evaluated using ${Data}_A$ dataset. Figure \ref{behavior} shows the detection performance of HUIHEN whose number of sessions is set to 50, and the number of user behaviors in the same session is a hyper-parameter. Figure \ref{session} shows the detection performance of HUIHEN whose number of user behaviors in the same session is set to 25 and the number of sessions is a hyper-parameter.
\par
\par
From the above two figures, we can see that trends of HUIHEN using different number of user behaviors in the same session and HUIHEN using different number of sessions have a certain similarity. At the beginning with the increase in the number of sessions and the increase in the number of user behaviors collected in the same session, both AUC and KS show an upward trend. This shows that more user historical behaviors the more benefit to the description of user behavior habits. User behavior habits and overdue rate of loan application are closely related, thus the more user historical behaviors the more helpful to accurately predict the overdue rate of users. However, when the number of sessions and the user behaviors in the same session exceed a threshold (the number of sessions is 50, and the number of user behaviors in the same session is 25), the curve of both AUC and KS begin to flatten. This is mainly because that excessively long data sequence may bring in more noise, and it can be found that the most user behavior sequences are within this threshold. Especially for the operation sequences in the same session, an excessively long window threshold will lead to more behavior filling (to feed features into the model, the operation behavior within the same session must be filled up to the same length), which may have a negative impact on the performance of the final model. Finally, 50 sessions and 25 user behaviors in the same session are just at the balance point.
\subsubsection{Evaluation of The Importance of Field Aware Network}
The importance of field aware network is evaluated on two datasets, Table \ref{table5} shows the detection performance of HUIHEN using field aware network and not using field aware network.
\begin{table}[htbp]
	\centering
	\caption{Detection Performance of HUIHEN Using Field Aware Network and Not Using Field Aware Network}\label{table5}
	\begin{tabular}{cccc}
		\hline
		dataset & method & AUC & KS\\
		\hline
		${Data}_A$ & Field-aware & 0.6887 & 0.2845\\
		${Data}_A$ & Field-aware-without-category & 0.6871 & 0.2843\\
		${Data}_B$ & Field-aware & 0.6632 & 0.2543\\
		${Data}_B$ & Field-aware-without-category & 0.6619 & 0.2535\\
		\hline
	\end{tabular}
\end{table}
\par
According to Table \ref{table5}, HUIHEN using field aware network obtains the AUC of 0.6887 on ${Data}_A$, which is 0.23\% higher than the HUIHEN not using field aware network. HUIHEN using field aware network obtains the AUC of 0.6632 on ${Data}_B$, which is 0.2\% higher than the HUIHEN not using field aware network. In addition, HUIHEN using field aware network obtains the higher KS than HUIHEN not using field aware network on two datasets. It can be concluded that the model for performing mapping transformation on input features obtains better detection performance. So it is meaningful to adopt the field aware network to better capture the intra-field information using those embedding representations inside the same field (the embedding representations in different fields contain different information), namely, the field aware network layer is necessary. 
\par
\begin{figure}[htbp]
\centering
\subfigure[before mapping.]{
\begin{minipage}{0.5\linewidth}
\centering
\includegraphics[scale=0.9]{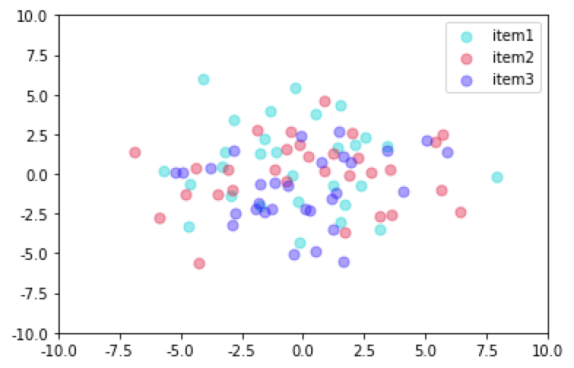}
%\caption{fig1}
\end{minipage}%
}%
\subfigure[after mapping.]{
\begin{minipage}{0.5\linewidth}
\centering
\includegraphics[scale=0.9]{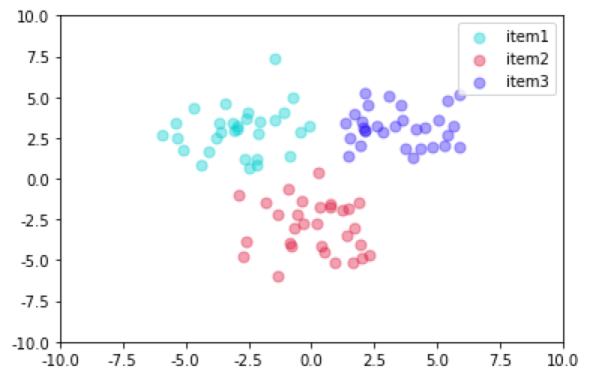}
%\caption{fig2}
\end{minipage}%
}%
\caption{the change of performing mapping transformation on items from three different fields.}\label{figure6}
\end{figure}

To further illustrate the effectiveness of the field aware network, we perform mapping transformation on items in three different fields. In order to facilitate visualization of data changes before and after mapping, the tsne \cite{ref53} algorithm is adopted to reduce the dimension of original item. The change that performs mapping transformation on items from three different fields is shown in Figure \ref{figure6}.
\par
According to Figure \ref{figure6}, before transforming features in different fields, all features are roughly in the same region, and there is no obvious boundary between features of different fields. After feature transformation, features in the same filed are mapped to the same region, features of different fields are mapped to different regions, and there is a clear boundary between them. Therefore, the field aware network is efficient.
\subsubsection{Discussion on The Universality and Rationality of The Teacher-Student Like Training Architecture}
The pre-training step in our training strategy adopts the teacher-student like training architecture. Section \ref{sy} has proved that the model using the teacher-student like training architecture obtains a better detection performance. In this section, to further discuss the universality and rationality of the teacher-student like training architecture, we conduct experiments on two more open source datasets about credit overdue prediction task, which are UCI dataset\footnotemark \footnotetext{\fontsize{5pt}{0.5em}\selectfont https://www.kaggle.com/uciml/default-of-credit-card-clients-dataset\par} and creditcard dataset\footnotemark \footnotetext{\fontsize{5pt}{0.5em}\selectfont https://www.kaggle.com/mlg-ulb/creditcardfraud/kernels\par}, respectively. The guidance model used in the experiment is Xgboost (LightGBM and Xgboost have similar guidance ability), and the target model is MLP (just like UPI Layer in Section \ref{upi}). AUC is used as the evaluation standard here, as AUC and KS have the same trend. The experimental results are shown in Table \ref{table6}.
\begin{table}[htbp]
	\centering
	\caption{The Universality and Rationality of The Teacher-Student Like Training Architecture}\label{table6}
	\begin{tabular}{cccc}
		\hline
		dataset & model & AUC\\
		\hline
		${Data}_A$ & MLP-guide-probability & 0.6705\\
		${Data}_A$ & MLP & 0.6614\\
		${Data}_B$ & MLP-guide-probability & 0.6514\\
		${Data}_B$ & MLP & 0.6489\\
		UCI & MLP-guide-probability & 0.7829\\
		UCI & MLP & 0.7692\\
		creditcard & MLP-guide-probability & 0.9776\\
		creditcard & MLP & 0.9690\\
		\hline
	\end{tabular}
\end{table}
\par
From Table \ref{table6}, MLP using teacher-student like training architecture obtains the AUC of 0.6705 on ${Data}_A$, which is 1.38\% higher than ordinary MLP (0.6614). MLP using teacher-student like training architecture obtains the AUC of 0.6514 on ${Data}_B$, which is 0.39\% higher than ordinary MLP (0.6489). MLP using teacher-student like training architecture obtains the AUC of 0.7829 on UCI, which is 1.78\% higher than ordinary MLP (0.7692). MLP using teacher-student like training architecture obtains the AUC of 0.9776 on credit, which is 0.89\% higher than ordinary MLP (0.9690). Therefore, the AUC of MLP model using teacher-student like training architecture has a significant improvement on all datasets, thus illustrating the universality of the teacher-student like training architecture proposed in this paper.
\section{Conclusion and Future Work}
As the application process of online mobile loan services need to be simple and fast which is not conducive to construct users'credit profile and predict the overdue  rate. If the risk of overdue cannot be effectively identified, it will bring huge economic losses to the bank. Therefore, how to use limited user information to calculate users’ credit loan overdue risks is a challenging task. In this paper, we leverage users' behavior sequences as a supplementary property, which can reflect users' intentions and habits that are helpful to predict the overdue rate. HUIHEN (Hierarchical User Intention-Habit Extract Network) are proposed to make full use of the information of user behavior sequences. First, we propose a strategy to embed the user behavior sequences, divide them into sessions, and introduce the field-aware network to extract the intra-field information of the behaviors. Then a hierarchical architecture is designed, in which the time-aware GRU and the user-item aware GRU are used to extract the users’ short-term intention and users’ long-term habits, respectively. The performance of HUIHEN consistently dominates the state-of-the-art methods, without increasing the complexity of the online application process and requiring users to fill in more information. This research has opened a new way of designing the credit decision model for the situation that contains much user behavior sequence data.
\par
For future work, as there is usually a time interval (usually 1-2 months) from the user's loan to judge whether the user is overdue, so many data collected during this period have no label information, especially at the beginning of the online serving time. How to make use of unsupervised algorithms such as word2vec to perform unsupervised pre-learning strategy to mine the valuable information from large amount unlabeled user behavior sequences data is well worth exploring in the future.




%\begin{thebibliography}{00}


%\bibitem{}

%\end{thebibliography}
\section{Acknowledgments}
We would like to thank Haojie Bai and Xiaozhong Kang for their help in the deployment of HUIHEN. We would also like to thank Lamei Zhao and Yilin Guo for the constructive suggestions and the anonymous reviewers for their insightful comments.
%\section*{References}
\bibliography{reference}
\end{document}